\documentclass{article}

\usepackage{microtype}
\usepackage{graphicx}
\usepackage{multirow}
\usepackage{booktabs}
\usepackage{subcaption}
\usepackage{booktabs} 
\usepackage{multicol}

\usepackage{hyperref}

\usepackage[accepted]{icml2025}

\usepackage{amsmath}
\usepackage{amssymb}
\usepackage{mathtools}
\usepackage{amsthm}

\usepackage[capitalize,noabbrev]{cleveref}

\definecolor{limegreen}{rgb}{0.2, 0.8, 0.2}
\definecolor{ered}{rgb}{0.72, 0.16, 0.2}
\definecolor{eblue}{rgb}{0.36, 0.36, 0.36}
\definecolor{epurple}{rgb}{0.8, 0., 0.8}
\definecolor{eorange}{rgb}{0.98, 0.3, 0.1}
\definecolor{running}{rgb}{0.05, 0.5, 0.95}
\definecolor{update}{rgb}{0.05, 0.5, 0.95}

\newcommand{\std}[1]{\textbf{\scriptsize\textcolor{eblue}{$\pm#1$}}} 

\theoremstyle{plain}

\theoremstyle{definition}

\theoremstyle{remark}

\usepackage[textsize=tiny]{todonotes}

\icmltitlerunning{Zero-Shot Model Search from Weights}

\begin{document}

\twocolumn[
\icmltitle{Can this Model Also Recognize Dogs? Zero-Shot Model Search from Weights}





\begin{icmlauthorlist}
\icmlauthor{Jonathan Kahana}{huji}
\icmlauthor{Or Nathan}{huji}
\icmlauthor{Eliahu Horwitz}{huji}
\icmlauthor{Yedid Hoshen}{huji}
\end{icmlauthorlist}

\icmlaffiliation{huji}{School of Computer Science and Engineering\\The Hebrew University of Jerusalem, Israel}
\icmlcorrespondingauthor{Jonathan Kahana}{jonathan.kahana@mail.huji.ac.il}

\icmlkeywords{Machine Learning, ICML}

\vskip 0.3in
]


%
\printAffiliationsAndNotice{}  

\begin{abstract}

With the increasing numbers of publicly available models, there are probably pretrained, online models for most tasks users require. However, current model search methods are rudimentary, essentially a text-based search in the documentation, thus users cannot find the relevant models. This paper presents ProbeLog, a method for retrieving classification models that can recognize a target concept, such as "Dog", without access to model metadata or training data. Differently from previous probing methods, ProbeLog computes a descriptor for each output dimension (logit) of each model, by observing its responses on a fixed set of inputs (probes). Our method supports both logit-based retrieval ("find more logits like this") and zero-shot, text-based retrieval ("find all logits corresponding to dogs"). As probing-based representations require multiple costly feedforward passes through the model, we develop a method, based on collaborative filtering, that reduces the cost of encoding repositories by $3\times$. We demonstrate that ProbeLog achieves high retrieval accuracy, both in real-world and fine-grained search tasks and is scalable to full-size repositories. \href{https://jonkahana.github.io/probelog/}{https://jonkahana.github.io/probelog}

\end{abstract}

\section{Introduction}
\label{sec:introduction}

\begin{figure}[t]
    \centering
    \includegraphics[width=0.7\linewidth]{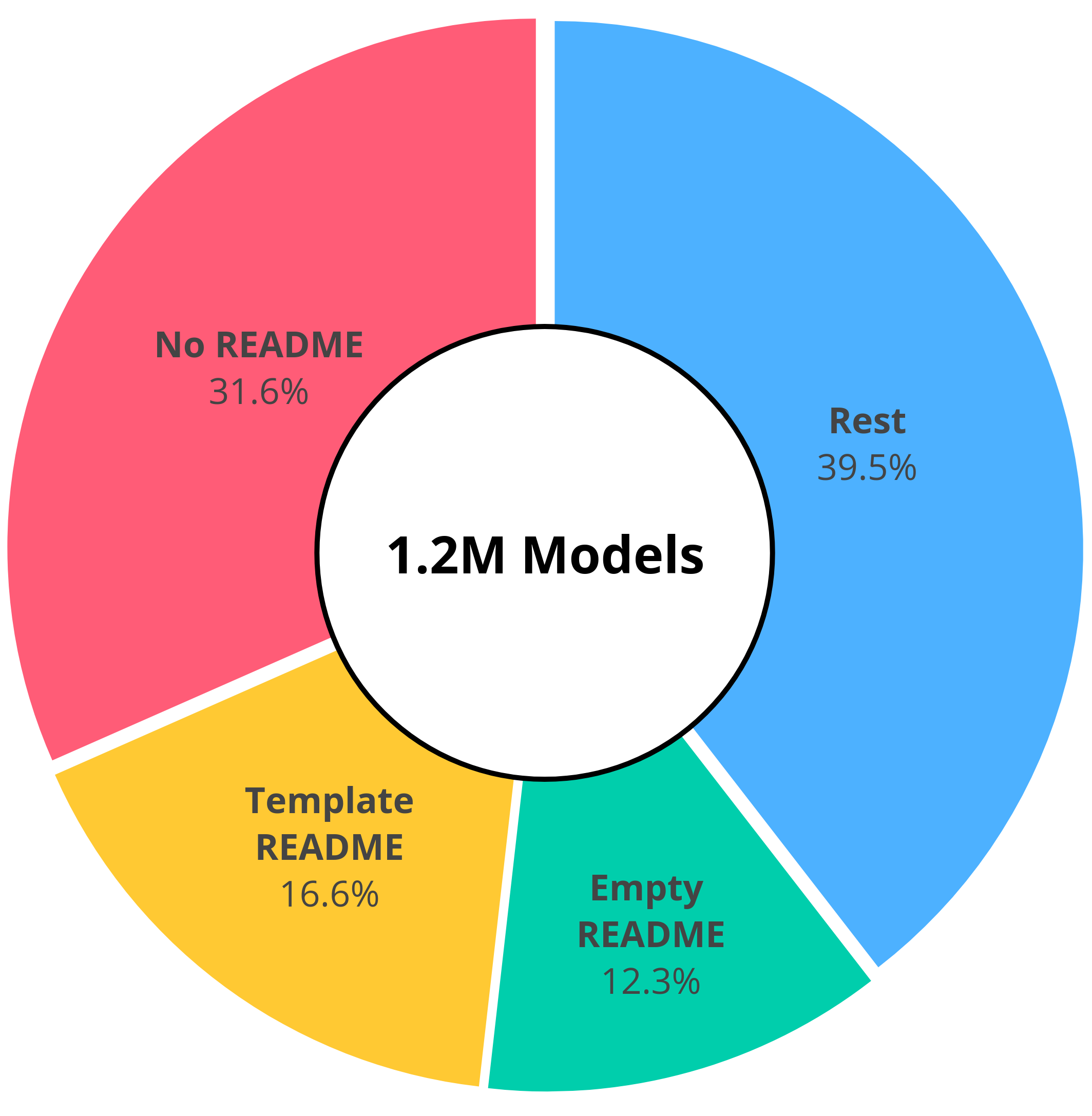}
    \caption{\textit{\textbf{Hugging Face Documentation.}} We analyze the model cards of $1.2M$ Hugging Face models. We discover that the majority of models are either undocumented or poorly documented.}
    \label{fig:hf_documentation}
\end{figure}

Neural networks have revolutionized fields like computer vision \citep{resnet,vit,yolo,clip,blip2,stable_diffusion} and natural language processing \citep{llama,bert,attentionisallyouneed}, becoming indispensable tools for many real-world classification tasks. However, their high training cost leaves users with two suboptimal options: i) invest heavily in computational resources for training or fine-tuning a model, ii) settle for a general-purpose model that may not suit their task. Now, imagine that instead, one could simply search online for the most accurate model for their specific task and use it directly without additional training. With the rise of large public model repositories, this is becoming feasible. For instance, Hugging Face, the largest existing model repository, hosts over 1 million models, with more than $100,000$ models added monthly. This significantly increases the likelihood of finding a suitable public model for most user tasks. The main challenge, however, lies in retrieving the right model for each task. While current model search methods \citep{hugginggpt,stylus} rely on provided metadata or text descriptions, most models in practice are either undocumented or have very limited descriptions (See Fig.~\ref{fig:hf_documentation}), which severely limits the ability of these search methods to retrieve suitable models.

\begin{figure*}[t]
    \centering
    \includegraphics[width=0.8\linewidth]{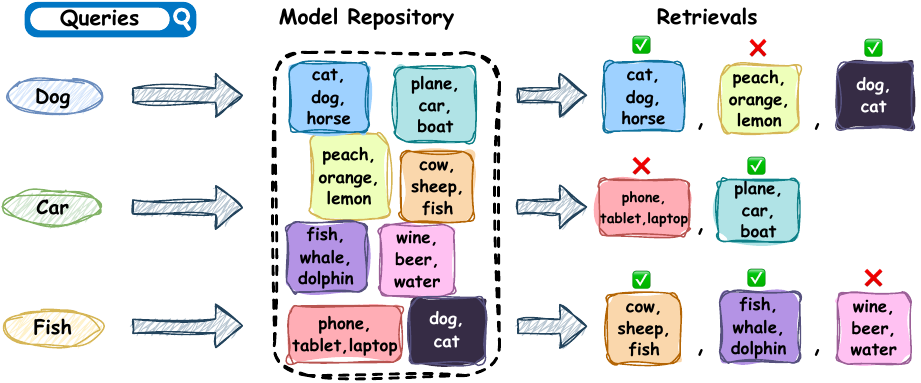}
    \caption{\textbf{\textit{Classification Model Search.}} We present a new task of Classification Model Search, where the goal is to find classifiers that can recognize a target concept. Concretely, given an input prompt, such as ``Dog'', we wish to retrieve all classifiers that one of their classes is ``Dog''. The search space is a large model repository, that contains many models and concepts to search from. The retrieved models can replace model training, increasing accuracy, reducing cost and environmental impact.}
    \label{fig:model_search}
\end{figure*}

We aim to search for new models based on their weights, without assuming access to their training data or metadata, as these are often unavailable. More precisely, the goal is to retrieve all classification models capable of recognizing a particular concept, such as ``Dog''. For a solution to be effective and practical, it must meet several requirements: i) identifying models that recognize the target concept regardless of the other concepts they can detect, ii) being invariant to model output class order, iii) scaling to large model repositories, and iv) supporting text-based search. Using a single representation to describe models is suboptimal for this task, as the target concept may only account for a small part of the representation. Model-level representations are often overly large, suffer from permutation variation and may be insensitive to the target concept. 

\begin{figure*}[t]
    \centering
    \includegraphics[width=0.78\linewidth]{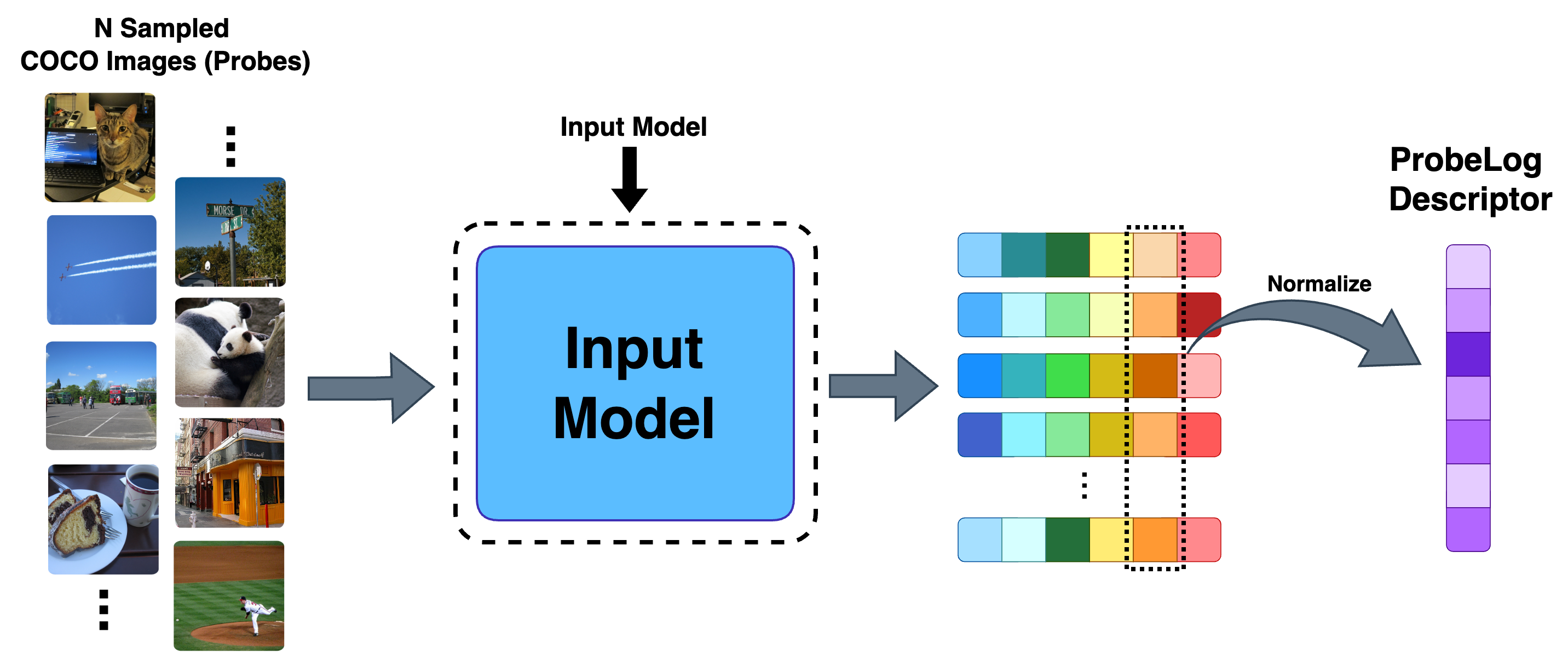}
    \caption{\textbf{\textit{ProbeLog Descriptors.}} Our method generates a descriptor for individual output dimensions (logits) of models. First, we sample and a set of inputs (e.g., from the COCO dataset), and fix them as our set of probes. Then, to create a new ProbeLog descriptor for a model logit, we feed the set of ordered probes nto the model and observe their outputs. Finally, we take all values of the logit we wish to represent, and normalize them. We use this representation to accurately retrieve model logits associated with similar concepts. In Fig.~\ref{fig:text_probelog}, we extend this idea to zero-shot concept descriptors.}
    \label{fig:probelog}
\end{figure*}

In this paper, we introduce \textit{ProbeLog}, a probing based logit-level descriptor especially designed for model search. Since our goal is to identify a 
functional property of the model (what it does), the descriptor is a functional representation \citep{non_interactive}, essentially it describes what the logit does. To compute the ProbeLog descriptor for a specific logit in a given model, we first query the model with a fixed set of pre-determined input samples (probes) and monitor its responses in the specific output dimension. By normalizing the response vector across all probes, we obtain the ProbeLog descriptor. Its dimension is equal to the number of probes. An illustration of ProbeLog descriptors is provided in Fig.~\ref{fig:probelog}. Crucially, unlike prior methods for analyzing neural network weights \citep{contentsearch}, our approach represents logits rather than the models, which are more suitable for search. 

ProbeLog representations enable searching by logit ("more like this"), but do not allow searching for unseen concepts ("find models that recognize 'dogs'"). Probably the most convenient way to achieve such zero-shot concept search is to incorporate text. Therefore, we propose to use a text alignment model (e.g., CLIP \cite{clip}) between the probes and target concept name to compute a zero-shot ProbeLog representation. After suitable domain normalization, this approach achieves accurate zero-shot search. To make ProbeLog practical for model search, we must address several questions. How does the choice of probes affect the representations? How can we choose effective probes suitable for various concepts? What similarity criteria should be used to compare between ProbeLog representations from two separate models? To answer these questions, in Sec.~\ref{sec:ablations} we conduct a thorough study of these questions and propose strategies to address them. As another core contribution, we present Collaborative Probing, a method to significantly reduce the cost of creating representations for a repository. Instead of probing all models with all probes, we only use a random selection of the probes for each model. We then complete the missing information with matrix-factorization based collaborative filtering. This results in greatly improved performance for low probe numbers.

 We showcase ProbeLog's effectiveness on two real-world datasets that we curate: one based on models that we train and the other containing models that we download from Hugging Face. Our method is scalable and can handle large models with high effectiveness and efficiency. It achieves high retrieval accuracy, reaching over $40\%$ top-1 accuracy when predicting whether a model can recognize an ImageNet \citep{deng2009imagenet} target concept from text. As the retrieval accuracy of a random method only scores $0.1\%$ (since there are $1,000$ possible classes), our method's performance is significant. Furthermore, we establish the strong performance of our Collaborative Probing approach.

Our main contributions are:
\begin{enumerate}
    \item Introducing ProbeLog, an effective logit-level model representation based on probing.
    \item Developing a method to extend this representation to zero-shot representations, enabling text-based model search. 
    \item Proposing Collaborative Probing to reduce the number of probes required for gallery encoding. 
\end{enumerate}

\section{Related Works}
\label{sec:related_works}

\subsection{Weight-Space Learning}

While neural networks can learn effective representations for many traditional data modalities, effective representations for neural networks are still work in progress. As a first step, \citet{statnn} proposed to observe simple statistics of weights, and  \citep{ke2017lightgbm} on them. Others proposed to encode the weights by modeling the connections between the neurons \citep{dws,inr2vec,sane,hyper_repr,eilertsen2020classifying,lim_lol,prem_neural_functionals,tran2024equivariant,functa,probex}. Recent methods \citep{neural_graphs,neural_functional_transformers,graph_meta,kalogeropoulos2024scale} model a network as a graph where every neuron is a node, and train permutation-equivariant architectures \citep{gilmer2017neural,kipf2016semi,attentionisallyouneed,relationalattention} on these graphs.
Probing is an alternative paradigm that encodes the network by observing its outputs to a fixed set of inputs (probes)  \citep{probegen,non_interactive,carlini2024stealing,tahan2024label,choshen2022start,neural_graphs,huang2024lg,dravid2023rosetta,bau2017network}. Differently from these approaches, we propose a probing-based method for zero-shot classification model search.

\subsection{Other Applications of Model Weights}

Learning on model weights has found many applications. Several approaches demonstrated advanced generation abilities using the weights \citep{dravid2024interpreting,erkocc2023hyperdiffusion,dravid2024interpreting,shah2023ziplora}, and others proposed to compress the weights to a smaller, more compact representation \citep{ha2016hypernetworks,nern,peebles2022learning}. A different line of research explored the relations between the weights for recovering the model graph \citep{mother,yax2025phylolm} or for merging \citep{yadav2024ties,gueta2023knowledge,izmailov2018averaging,wortsman2022model,rame2023model}. Recently, a few works proposed to recover the exact black-boxed weights \citep{spectral_detuning,carlini2024stealing} by having access to their fine-tuned versions or to an API. Finally, some relevant works search for new adapters for generative models \citep{hugginggpt,stylus,contentsearch}, however these approaches either rely on available metadata or tailored for generative models. Here, we propose an approach to search for new discriminative models which are capable of detecting a specific concept among other unrelated concepts seen in training time.

\section{Background and Motivation}

\subsection{Problem Definition: Model Search} 
We assume a model repository composed of $m$ classifiers, $f_1, f_2, ..., f_m$. Each classifier $f_i$ can have multiple output dimensions (logits), and each corresponding to an unknown concept $c_{i,j}$. The user then inputs a text prompt containing some query concept, $c_q$, they wish to search for. Finally, the goal is to return a model $f_i$ such that one of its classes matches the query concept. Formally, the set of all valid retrieval models, $R(c_q)$, is defined as:
\begin{equation}
    R(c_q) = \{f_i~~|~~\exists~j~~s.t.~~c_{i,j} = c_q\}
\end{equation}
As mentioned above, the retrieval algorithm does not know the class concepts of each model. We assume access to them solely for evaluation purposes. 

\subsection{The Challenge} While a trivial solution is to create model-level representations, this idea encounters serious setbacks. First, representing models by their weights is difficult and computationally expensive due to their high dimensionality and complex symmetries \citep{neural_graphs,probegen}. Second, encoding an entire model is not suitable for functionality-based search. To illustrate, consider a classifier that separates between ``Dog'',``Cat" and another one for ``Dog'' and ``Lion''. Despite both including the target concept ``Dog'', each of them will have a different encoding. Moreover, even classifiers with identical classes that are ordered differently (``Dog''-``Cat'' vs. ``Cat''-``Dog'') may produce distinct representations. To overcome this limitation and ensure invariance to other detected classes and their order, we propose to have a separate descriptor (representation) for each output dimension of each model.

\subsection{Real Models are Poorly Documented}
The existing solution for model search is text-based search in the user uploaded documentation. To understand the effectiveness of this solution, we explore the level of documentation of models in \href{https://huggingface.co/}{Hugging Face}, the largest model repository. For that, we analyzed all $1.2M$ model cards. As shown in Fig.~\ref{fig:hf_documentation}, over $30\%$ of all models have no model card at all. Moreover, there are another $28.9\%$ of model cards that are either empty or include an empty automatic template with no information. The remaining $40\%$ of model cards may include some information, however we cannot determine exactly how many of them include relevant information about the training data. As most models are poorly documented we conclude that searching models by weights alone is a practical and useful setting.

\begin{figure}[t]
    \centering
    \begin{tabular}{ccc}
        \begin{minipage}{0.135\textwidth}
            \centering
            \includegraphics[width=\textwidth]{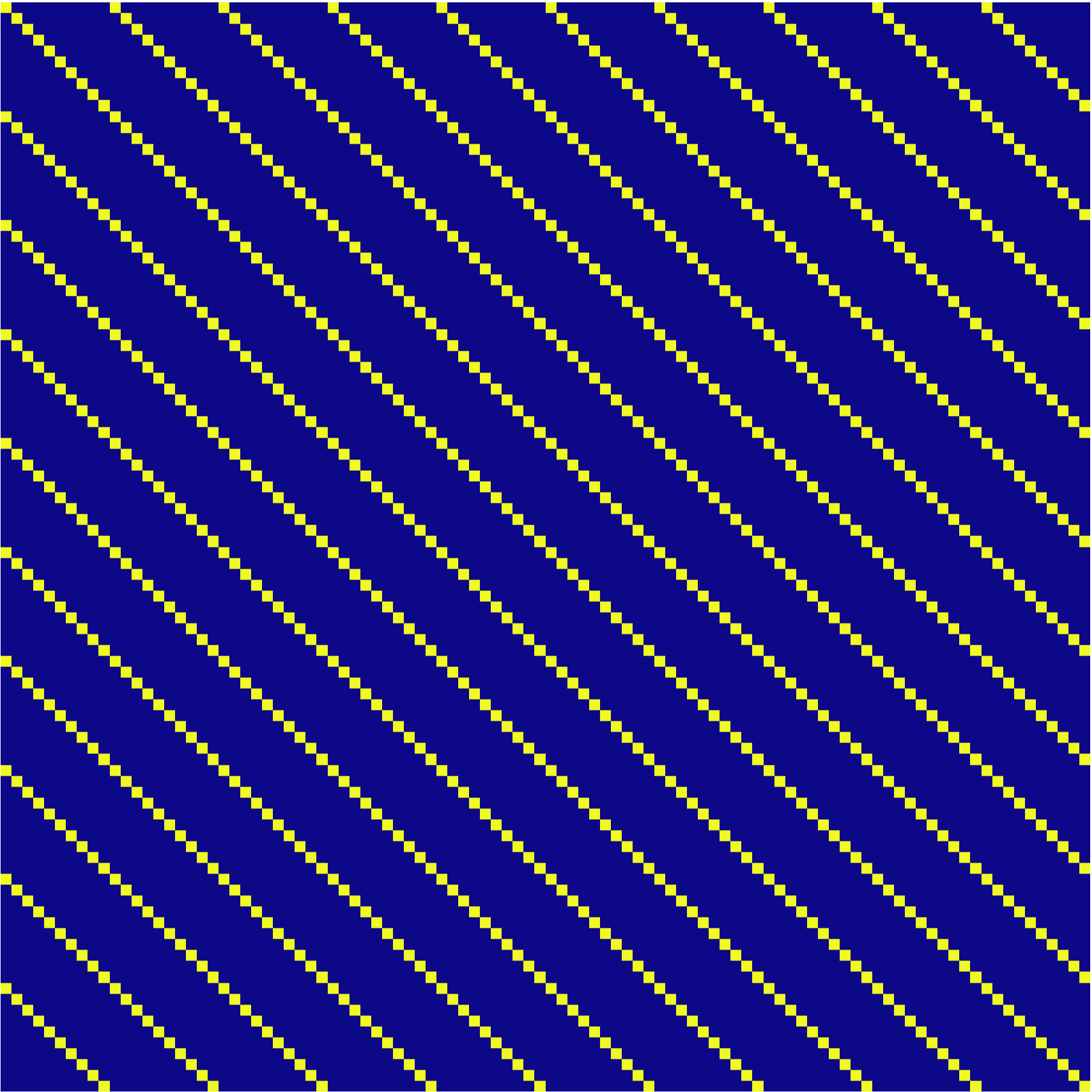}
            \subcaption{\textbf{\textit{GT}}}
            \label{fig:cifar_gt}
        \end{minipage} &
        
        \begin{minipage}{0.135\textwidth}
            \centering
            \includegraphics[width=\textwidth]{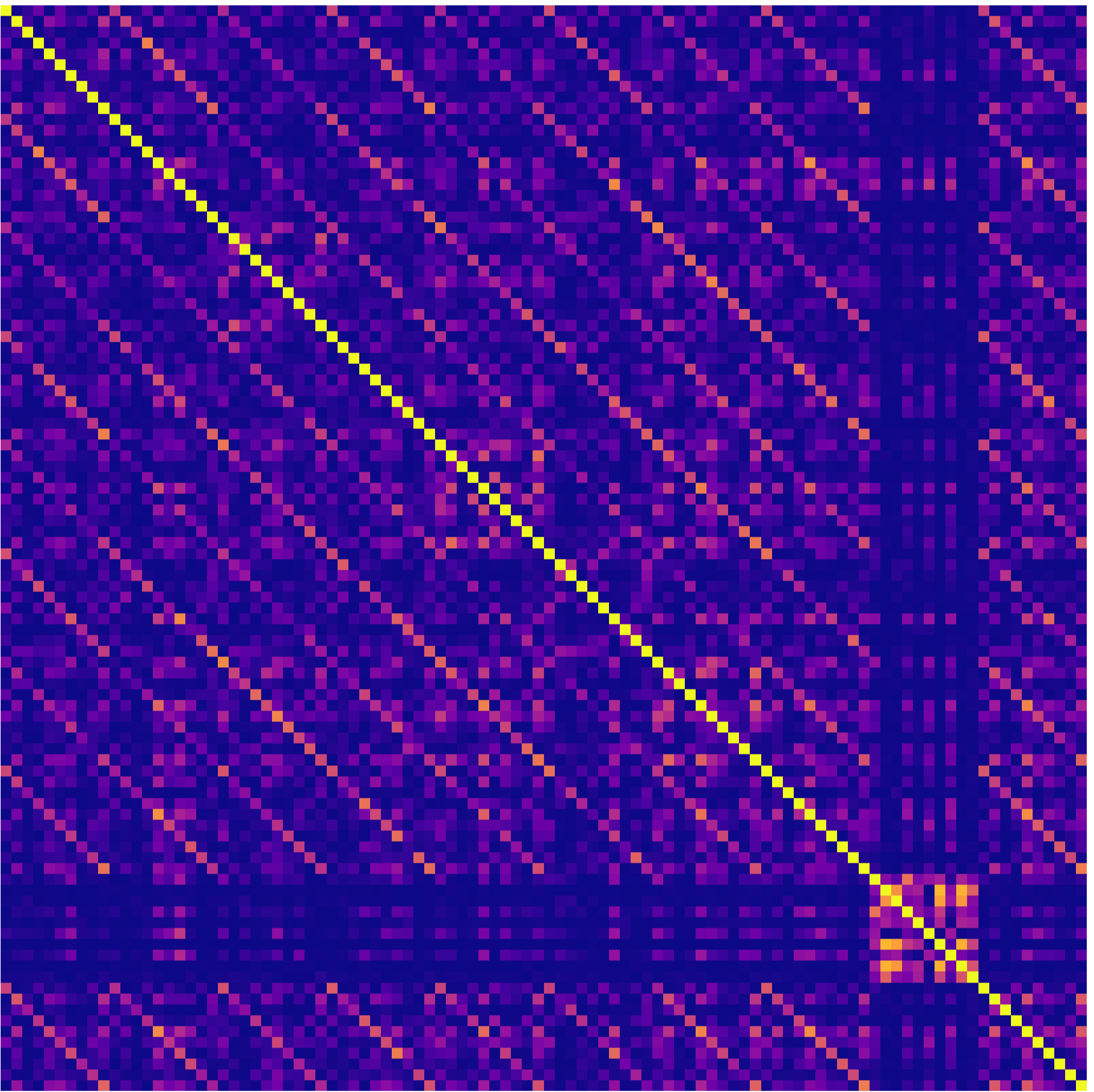}
            \subcaption{\textbf{\textit{Out-of-Dist.}}}
            \label{fig:cifar_coco_affinities}
        \end{minipage} &
        
        \begin{minipage}{0.135\textwidth}
            \centering
            \includegraphics[width=\textwidth]{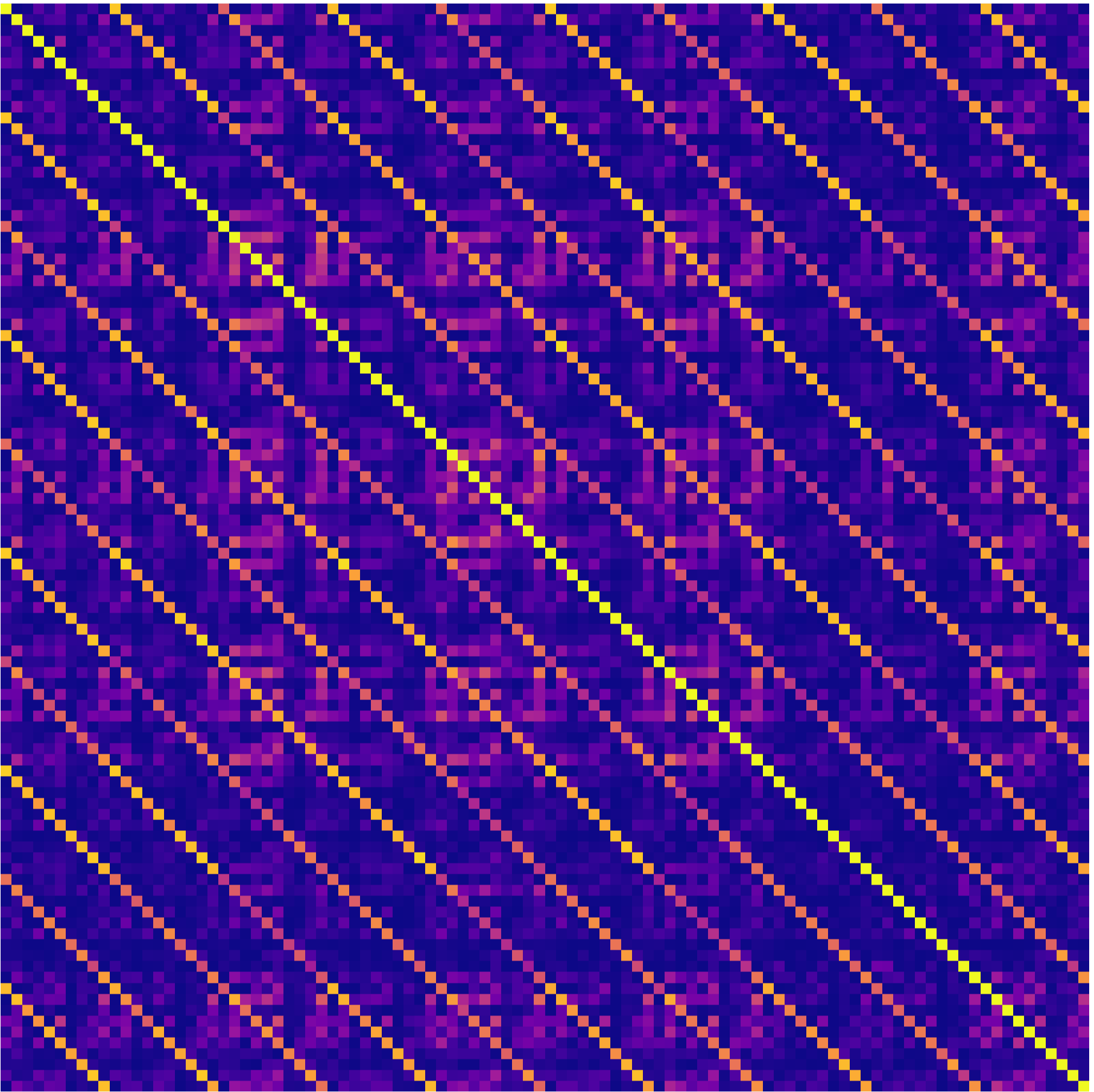}
            \subcaption{\textbf{\textit{In-Dist.}}}
            \label{fig:cifar_affinities_cifar_probes}
        \end{minipage}
    \end{tabular}
    \caption{\textbf{\textit{CIFAR10 Logit Similarities.}}(a) Ground truth label. (b) ProbeLog representations using $1,000$ out-of-distribution COCO image probes. (c) ProbeLog representations using $1,000$ in-distribution CIFAR10 image probes. Both find meaningful similarities, although in-distribution probes work better.}
    \label{fig:affinities}
    \vspace{-3mm}
\end{figure}

\section{Method}
\label{sec:probelog}

\subsection{ProbeLog: Logit-Level Descriptors}
\label{sec:descriptors}
Our objective is to accurately and efficiently find relevant models in a large repository that can recognize a target concept, e.g., ``Dog''. Instead of using a single representation for the entire model, we represent each model output (logit) separately. Our method for extracting logit descriptors first presents each model with a set of $n$ ordered, fixed input samples (probes). Intuitively, these are a set of standardized questions that we ask the model. In practice, we compose the list of probes by randomly sampling images (without replacement) from an image dataset. For generality, most of our results use the COCO dataset \citep{coco} which is highly diverse but also out-of-distribution to our models. We investigate the choice of probe dataset in Sec.~\ref{sec:ablations}. We input each probe $x_j$ into the model $f$, obtaining the output $f(x_j)[i]$ for the model's $i^{th}$ logit. We define the ProbeLog descriptor for logit $i$ of model $f$ as the responses of all probes at this logit:
\begin{equation}
    \phi(f,i) = [f(x_1)[i],f(x_2)[i],\cdots,f(x_n)[i]]
\end{equation}
Fig.~\ref{fig:probelog} presents an overview of ProbeLog extraction.

To validate that logit responses to probes provide an effective description of the semantic function, we present a simple experiment. We take $10$ different ViT foundation models, each trained via a different procedure and fine-tune them on the CIFAR10 \citep{cifar10} classification task (classifying small images into one of $10$ object categories). We randomly sample $1,000$ ImageNet \citep{deng2009imagenet} images as probes and run them through the model, computing the ProbeLog description of each logit in each model ($100$ in total). We then compute the correlation between all pairs of logits, and present the correlation matrix in Fig.~\ref{fig:affinities}. We observe that logit responses to probes are mostly correlated to those of logits with a matching semantic concept instead of, for example, logits from the same model.

\subsection{A Discrepancy Measure for Logit-Level Descriptors}
\label{sec:similarity_metric}

For downstream tasks such as model retrieval, we need to compute the discrepancy between pairs of logit-level ProbeLog descriptors. However, naive metrics such as Euclidean or correlation yield subpar results. We hypothesize that models are only reliable for probes they are confident about, while their responses exhibit high variance for the others. 
To mitigate this phenomenon, we propose focusing only on probes for which the query logit has high confidence about.  
We introduce an asymmetric discrepancy measure, specifically designed for logit-level comparisons. Given a query logit descriptor, $\phi$, we sort its values (probe responses) from highest to lowest. Let $\textit{a} = [a_1, a_2, \dots, a_n]$ be the indices of the sorted entries in descending order. We then reorder all gallery descriptors using the same index sequence $\textit{a}$. Lastly, we compute the discrepancy between the query and each of the gallery descriptors by measuring the difference (in $L_2$) only over the top k probe entries of the sorted descriptors: 

\begin{equation}
d(\phi, \phi') = \sqrt{\sum_{i=1}^k \left(\phi_{a_i} - \phi'_{a_i}\right)^2}    
\end{equation}

where $d(\phi, \phi')$ is the discrepancy between the query descriptor $\phi$ and a gallery descriptor $\phi'$. In Sec.~\ref{sec:ablations} we show the importance of these design choices.

\begin{figure}[t]
    \centering
    \includegraphics[width=0.97\linewidth]{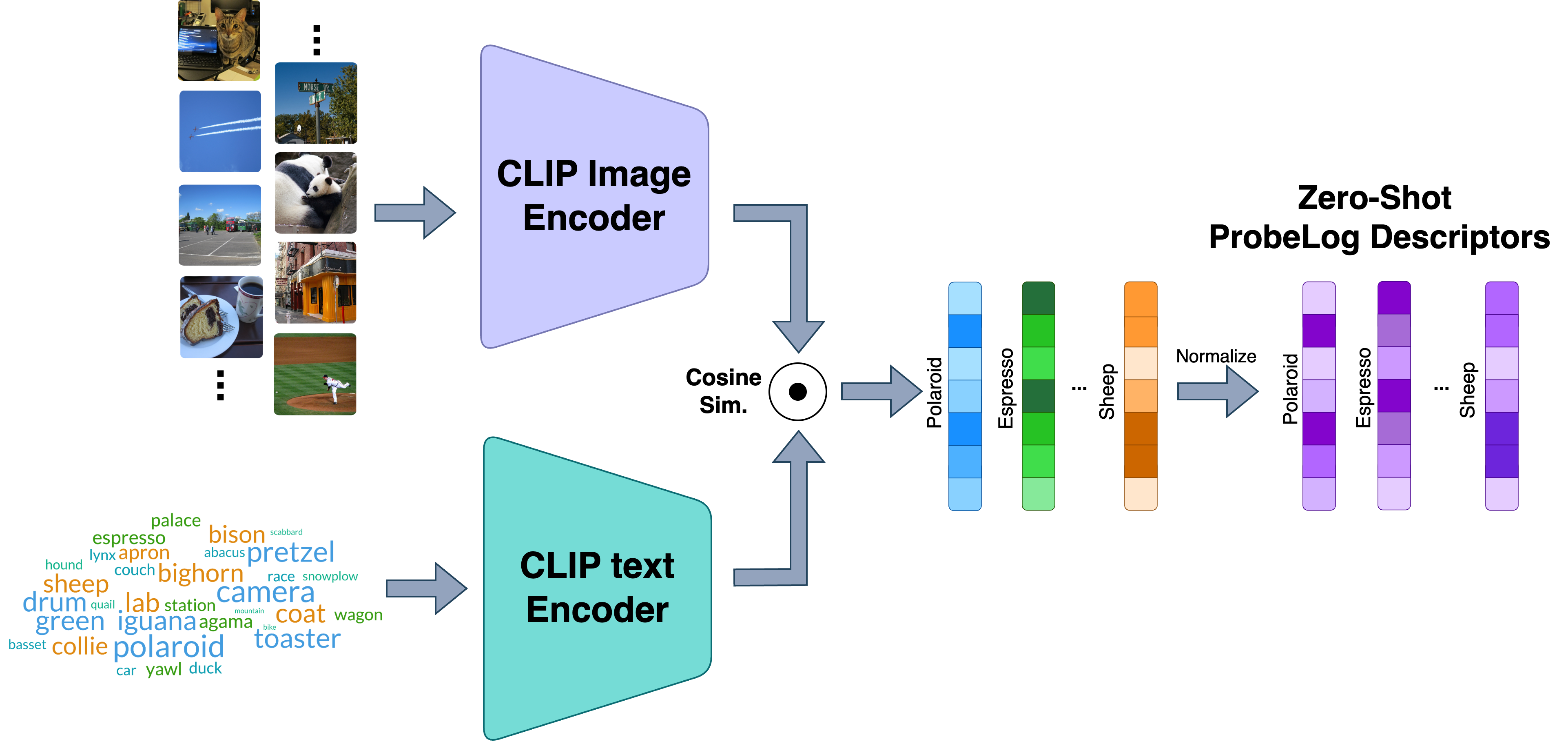}
    \caption{\textit{\textbf{Text-Aligned ProbeLog Representation.}} We present a method to create ProbeLog-like representations for text prompts. We encode and store each of our ordered probes using the CLIP image encoder. At inference time, we embed the target text prompt, and compute its similarity with respect to the stored probe representations. We demonstrate that by normalizing this zero-shot ProbeLog descriptor, we can effectively search descriptors of real model logits, accurately retrieving similar concepts.}
    \label{fig:text_probelog}
    \vspace{-3mm}
\end{figure}

\subsection{Text-Aligned ProbeLog Descriptors}
\label{sec:zero_shot_descriptors}

The previous sections provided a way to search by logit, essentially finding similar logits to an existing one. This limits its applicability as it assumes the user already has such model. In this section, we extend our method to searching by text, thus allowing the user to search for concepts without already having such model, making it zero-shot. To do so, we present a method for generating ProbeLog descriptors from text alone. We use a multimodal text alignment model. For example, when the inputs are images, we choose CLIP, a joint text-image embeddings model. We use the multimodal model to extract embeddings from each probe $\alpha_i$ as well as from a user description $\alpha_{text}$ of the target concept. We define the zero-shot ProbeLog descriptor of the target concept as the vector of dot products between the embeddings of each probe and that of the target text:
\begin{equation}
    \phi_{text} = [\alpha_i \cdot \alpha_{text}, \alpha_2 \cdot \alpha_{text}, \cdots, \alpha_n \cdot \alpha_{text}]
\end{equation}

Using our discrepancy measure between the logit and the zero-shot ProbeLog descriptors does not achieve good results as their numerical values are in different scales. To reduce this domain gap, we normalize each descriptor by its mean and standard deviation. The normalized ProbeLog descriptor is:
\begin{equation}
    \phi(f,i) \leftarrow \frac{\phi(f,i) - \mu_{f,i}}{\sigma_{f,i}}
\end{equation}
, where $\mu_{f,i}$ and  $\sigma_{f,i}$ indicate the mean and standard deviation of $\phi(f,i)$ respectively. We illustrate the creation of our zero-shot ProbeLog descriptors in Fig.~\ref{fig:text_probelog}.

\begin{figure*}[t]
    \centering
    \includegraphics[width=0.76\linewidth]{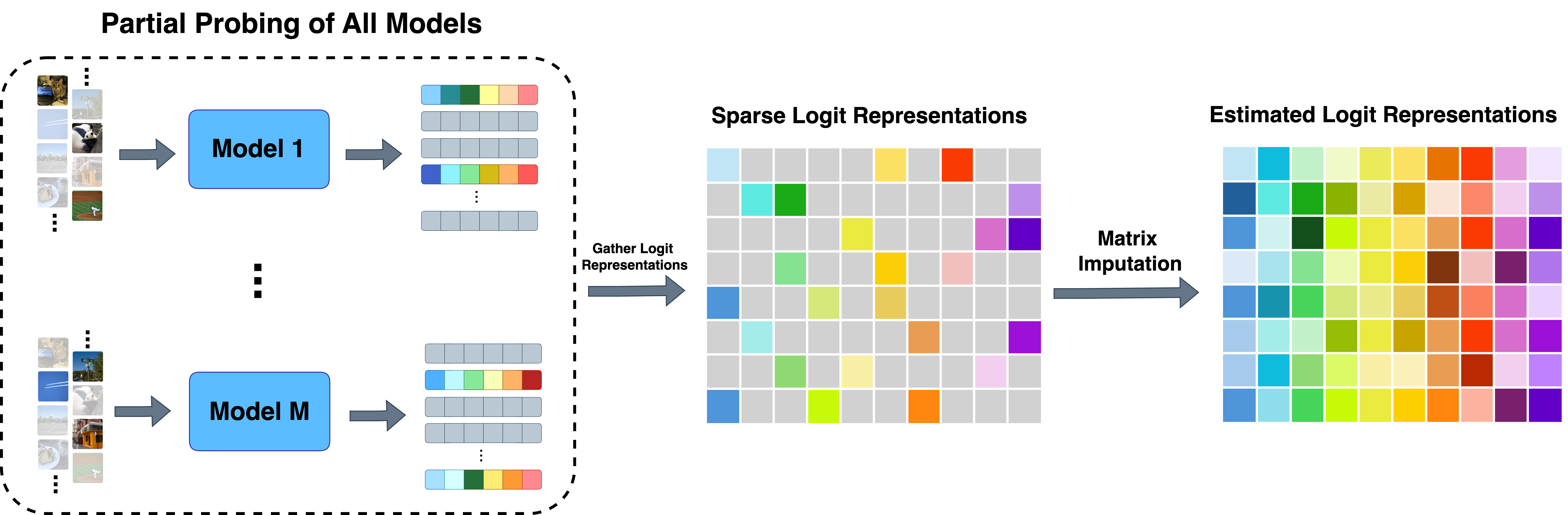}
    \caption{\textit{\textbf{Collaborative Probing.}} We pass a random subset of probes through each model in the repository to obtain partial logit representations. By performing factorization based matrix imputation we can complete the missing information. This saves a substantial part of the computational resources needed to build our repository's logit descriptors gallery.}
    \label{fig:collaborative_probing_diagram}
\end{figure*}

\subsection{Collaborative Probing}

Creating the ProbeLog representations for an entire model repository can be very costly, as it requires computing many forward passes for millions of models. Reducing this number of probes is critical for making the method feasible. Therefore, we propose Collaborative Probing. For each model, we randomly sample $p\%$ of the probes, and compute the ProbeLog representation just on these probes, masking out the entries for the other probes. We can therefore describe the ProbeLog descriptors for the logits of all models in the repository as a sparse matrix $X$, with $1-p\%$ of entries missing, where $X_{i,j}$ is the response of logit $i$ to probe $j$. The core idea is to use missing data imputation methods to complete this matrix, thus cheaply computing the full ProbeLog representations while actually probing each model with only a small fraction of the probes.

To complete the matrix $X$, we use the truncated SVD algorithm \cite{koren2009matrix} that famously won the Netflix prize for movie recommendation. The idea is to decompose matrix $X$ into low-rank matrices $U,V$ such that:
\begin{equation}
    U^*,V^* = arg\min_{U,V} |(U^TV - X) \odot M|^2
\end{equation}
where $M$ is the mask matrix that has all ones expect for zeros for masked entries of $X$. We solve this optimization problem using iterative optimization. This involves alternating between fixing $U$ while optimizing $V$ and vice versa until convergence. By the end of optimization, we compute $\tilde{X} = U^TV$, the completed matrix. Computing the zero-shot ProbeLog embedding does not require any modification, as the probe embeddings can be cached. At inference time, the text embedding requires a single forward pass, and the zero-shot ProbeLog descriptor requires a single matrix vector multiplication. The retrieval then proceeds normally.

\begin{table*}[t]
    \caption{\textbf{\textit{Retrieval Results.}} We evaluate the Top-1 and Top-5 retrieval accuracies of our method and the baselines for search-by-logit and search-by-text. All methods use COCO images as probes. For a fair comparison, all experiments are performed with $4,000$ probes.}
    \vspace{0.2cm}
    \centering
    \begin{tabular}{c|cccccc}
    Retrieval & Method & INet $\rightarrow$ INet & INet $\rightarrow$ HF & HF $\rightarrow$ INet & text $\rightarrow$ HF & text $\rightarrow$ INet \\
    \toprule
    \multirow{2}{*}{Top-1} & Full Query & 59.9\%\std{0.2} & 14.8\%\std{0.1} & 15.3\%\std{0.8} & 22.6\%\std{0.5} & 16.9\%\std{0.2} \\
    \multirow{2}{*}{Accuracy} & Model-Level & 0\%\std{0.} & 13.9\%\std{1.0} & 21.0\%\std{1.8} & 17.8\%\std{1.5} & 0\%\std{0.0} \\
    & \textbf{Ours (ProbeLog)} & 72.8\%\std{0.2} & 26.1\%\std{0.8} & 40.6\%\std{0.3} & 34.0\%\std{1.5} & 43.8\%\std{1.1} \\
    \midrule
    \multirow{2}{*}{Top-5} & Full Query & 82.8\%\std{0.1} & 31.5\%\std{0.1} & 19.7\%\std{0.8} & 38.6\%\std{1.1} & 22.8\%\std{0.2} \\
    \multirow{2}{*}{Accuracy} & Model-Level & 0\%\std{0.} & 34.6\%\std{0.6} & 51.6\%\std{2.0} & 38.8\%\std{1.8} & 0\%\std{0.0} \\
    & \textbf{Ours (ProbeLog)} & 92.6\%\std{0.1} & 43.6\%\std{0.5} & 58.6\%\std{0.9} & 53.7\%\std{1.9} & 68.0\%\std{0.6} \\
    \end{tabular}
    \label{tab:main_results}
\end{table*}

\section{Experiments}

\subsection{Experimental Setting}

\textbf{Datasets.} As there are no suitable existing datasets for model search that include ground-truth data, we created $2$ new ones, INet-Hub and HF-Hub. For each model in the INet-Hub, we sample a subset of ImageNet classes, a model architecture and foundation model initialization checkpoint. We then train the model on the selected data. The final dataset consists of $1,500$ models, making a total of more than $85,000$ logits (consisting of $1000$ unique fine-grained concepts). For more details, see App.~\ref{app:inet_hub}. Our second hub,  HF-Hub, is a set of $71$ real-world models ($400$ logits) downloaded manually from HuggingFace. As these data were created by real Hugging Face users, the concepts names might partially overlap (e.g., "Apple" vs. "Apples"). We therefore manually label the allowed retrievals with respect to this dataset and to ImageNet classes (see App.~\ref{app:hf_hub}). 

\textbf{Baselines.} We test our retrieval algorithm against two baselines: (i) model-level, and (ii) direct logit comparison. The model-level approach averages all ProbeLog descriptors of the model's logits, and searches for a similar logit descriptor to that model-level representation. The logit-level baseline does not use our discrepancy metric, but computes the Euclidean distance between a pair of logit representations.

\textbf{Metrics.} We evaluate the retrieval performance using standard metrics: top-k accuracy and precision (with $k \in [1,5]$). Top-k accuracy measures the percentage of target logits that had a relevant result in any of their top-k retrieved logits. Top-k precision measures the percentage of all top-k retrievals across all target concepts that were relevant.

\begin{table*}[t]
    \caption{\textbf{\textit{Dataset Ablations.}} We compare both real and synthetic probe distributions. While distributions closer to the model’s training data lead to better results, even out-of-distribution probes sampled from the COCO dataset retrieve relevant logits with high accuracy.}
    \vspace{0.2cm}
    \centering
    \begin{tabular}{cccc|ccc}
        & \multicolumn{3}{c}{Top-1 Accuracy} & \multicolumn{3}{c}{Top-5 Accuracy} \\
        Method & HF $\rightarrow$ INet & text $\rightarrow$ HF & text $\rightarrow$ INet & HF $\rightarrow$ INet & text $\rightarrow$ HF & text $\rightarrow$ INet \\
        \toprule
        Dead-Leaves & 1.3\%\std{0.7} & 1.6\%\std{1.4} & 1.0\%\std{0.2} & 5.9\%\std{1.6} & 6.8\%\std{1.3} & 3.8\%\std{0.2} \\
        Stable-Diffusion & 51.4\%\std{1.0} & 36.9\%\std{0.9} & 47.0\%\std{0.6} & 69.8\%\std{0.9} & 56.2\%\std{0.9} & 73.3\%\std{0.9} \\
        ImageNet & 57.8\%\std{1.3} & 33.1\%\std{1.2} & 55.4\%\std{1.1} & 71.4\%\std{1.3} & 55.1\%\std{0.9} & 80.4\%\std{0.9} \\
        COCO & 40.6\%\std{0.3} & 34.0\%\std{1.5} & 43.8\%\std{1.1} & 58.6\%\std{0.9} & 53.7\%\std{1.9} & 68.0\%\std{0.6} \\
    \end{tabular}
    \label{tab:dataset}
\end{table*}

\subsection{Model Search Results}
\label{sec:main_results}

We evaluate our method on 3 scenarios and present the results in  Tab.~\ref{tab:main_results}. 
Collaborative Probing is evaluated separately in Sec.~\ref{sec:cp}. Here, we report top-1/5 accuracies, for top-5 precision results see App.~\ref{app:additional_results}. 

In the first scenario, we evaluate our performance when target models come from the same distribution as the repository models. To test this, we split the INet-Hub into 2 distinct subsets, and evaluate the retrieval performance. In this setting, ProbeLog achieves excellent accuracy, with a top-1 accuracy of $70\%$ i.e., more than two thirds of target logits have the correct concept as their top retrieval result.

The second scenario is the more difficult case, where the queries are out-of-distribution to the repository. To test this, we search for real model logits (HF-Hub) in the INet-Hub and vice versa. This is especially difficult as the INet-Hub contains logits corresponding to ImageNet classes that are quite fine-grained. Still, ProbeLog obtains top-1 retrieval accuracy of $40.6\%$ in the $HF \rightarrow INet$ task, compared to both baselines which are at $21\%$.

In the search-by-text evaluation, we search for the closest retrievals to a zero-shot text descriptor in either the INet-Hub or the HF-Hub. We can see that in both cases, our approach greatly exceeds the baselines, reaching an impressive top-1 accuracy of $43.8\%$ on the INet-Hub. Moreover, when tested on the HF-Hub we can see that our method generalizes to real-world models, as it finds suitable matches for more than a third of the queries in the first search result, and for more than half of the queries within the first $5$ retrievals. This shows that while simple, our approach can generalizes to a real-world scenarios where user models are searched for using just a simple text prompt.

\subsection{Collaborative Probing}
\label{sec:cp}

\begin{figure}[t]
    \centering
    \includegraphics[width=\linewidth]{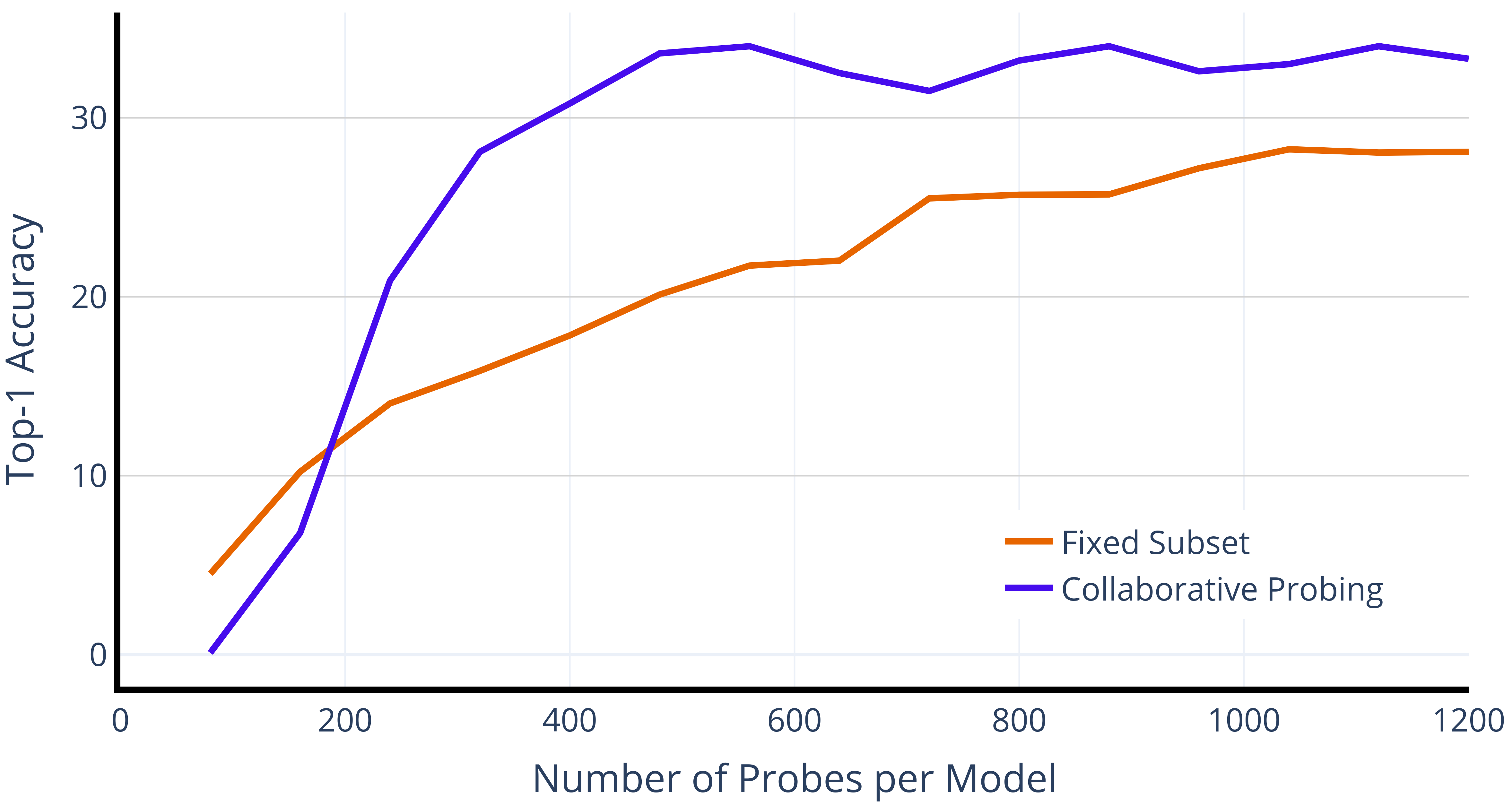}
    \caption{\textit{\textbf{Collaborative Probing.}} We test our method using collaborative probing on the text $\rightarrow$ INet-Hub retrieval task. While the full size of the dataset is $8,000$ COCO probes, we show cases where each model is probed by less than $15\%$ of these probes. We can see that for the limited probe regime, collaborative probing can improve accuracy by as much as $2\times$.}
    \label{fig:collaborative_probing}
    \vspace{-5mm}
\end{figure}

We compare Collaborative Probing, sampling a number of randomly selected probes for each model against simply using the same probes for all models. The results are presented in Fig.~\ref{fig:collaborative_probing}. While our collaborative probing technique requires around $400$ probes per model to be effective, it can then substantially improve probing efficiency. Specifically, it reaches similar results as the standard approach with less than a third the number of probes. For example, having just $4\%$ of all probes per model, is just as good as probing all models with $15\%$ of all probes. This highlights the potential of our collaborative probing technique to significantly improve the efficiency of our search approach.

\subsection{Ablation Studies}
\label{sec:ablations}

\paragraph{How to select the probe distribution?} We showed (Sec.~\ref{sec:main_results}) that ProbeLog can generalize to real-world scenarios. Here, we conduct an ablation study, to test the effect of sampling probes from different distributions: (i) Dead-Leaves \citep{baradad2021learning,lee2001occlusion}: a very coarse, hand-crafted generative model. (ii) ImageNet images. (iii) StableDiffusion \citep{stable_diffusion} samples using prompts of ImageNet-21K objects. (iv) COCO Images. Results, shown in Tab.~\ref{tab:dataset}, demonstrate a consistent pattern: probes sampled from distributions that are closer to the target concept obtain more accurate retrievals. However, we note that even quite different probe distribution can yield high retrieval accuracies. E.g., even though COCO images are typically of scenes rather than objects, they are effective probes, reaching a top-5 accuracy of more than $60\%$ when searching the INet-Hub by text. These results show that defining a general set of probes, which can retrieve a wide range of concepts is feasible. However, if a prior knowledge about the distribution of target concepts exists, then it is better to select in-distribution probes.

\begin{table*}[t]
    \caption{\textbf{\textit{Logit Discrepancy Ablations.}} Our evaluation reveals: i) normalizing logit descriptors is necessary for accurate retrieval, especially for search-by-text. ii) choosing the most confident probes of the query logit is crucial, no other approach achieved comparable accuracy.}
    \vspace{0.2cm}
    \centering
    \begin{tabular}{cccc|ccc}
        & \multicolumn{3}{c}{Top-1 Accuracy} & \multicolumn{3}{c}{Top-5 Accuracy} \\
        Selected Probes & HF$\rightarrow$INet & text$\rightarrow$HF & text$\rightarrow$INet & HF$\rightarrow$INet & text$\rightarrow$HF & text$\rightarrow$INet \\
        \toprule
        Top-$k$ + No Norm. & 1.9\%\std{0.4} & 0.1\%\std{0.1} & 0\%\std{0.0} & 5.0\%\std{0.9} & 0.5\%\std{0.2} & 0.6\%\std{0.1} \\ 
        Bottom-$k$ & 0.8\%\std{0.3} & 1.3\%\std{0.9} & 2.3\%\std{0.5} & 1.2\%\std{0.3} & 6.8\%\std{0.5} & 7.9\%\std{0.9} \\
        Random & 8.6\%\std{1.2} & 2.5\%\std{1.4} & 6.3\%\std{0.5} & 16.2\%\std{1.5} & 8.6\%\std{1.2} & 16.7\%\std{0.7} \\
        Quantiles & 7.7\%\std{2.0} & 5.9\%\std{1.0} & 5.8\%\std{0.5} & 17.9\%\std{3.8} & 15.3\%\std{1.8} & 16.4\%\std{1.7} \\
        All & 15.3\%\std{0.8} & 22.6\%\std{0.5} & 16.9\%\std{0.2} & 19.7\%\std{0.8} & 38.6\%\std{1.1} & 22.8\%\std{0.2} \\
        \textbf{Top-$k$ (Ours)} & 40.6\%\std{0.3} & 34.0\%\std{1.5} & 43.8\%\std{1.1} & 58.6\%\std{0.9} & 53.7\%\std{1.9} & 68.0\%\std{0.6} \\
    \end{tabular}
    \label{tab:sim_metric}
\end{table*}

\paragraph{Which probes should be in the discrepancy metric?} We proposed a discrepancy metric that compares the query and retrieved logits only on the probes that the query logit obtained large values on (Sec. \ref{sec:descriptors}). We ablate this choice of metric, comparing to several other probe selection criteria: lowest value probes, random sampling, uniform quantile sampling, highest value probes without normalization, and using all probes. The results, presented in Tab.~\ref{tab:sim_metric}, show that selecting the highest valued probes of the query logit is crucial for successful retrieval. We believe this is because logit values tend to be noisy, and highly confident values should be more consistent across logits of the same concept.

\begin{figure}[t]
    \centering
    \includegraphics[width=\linewidth]{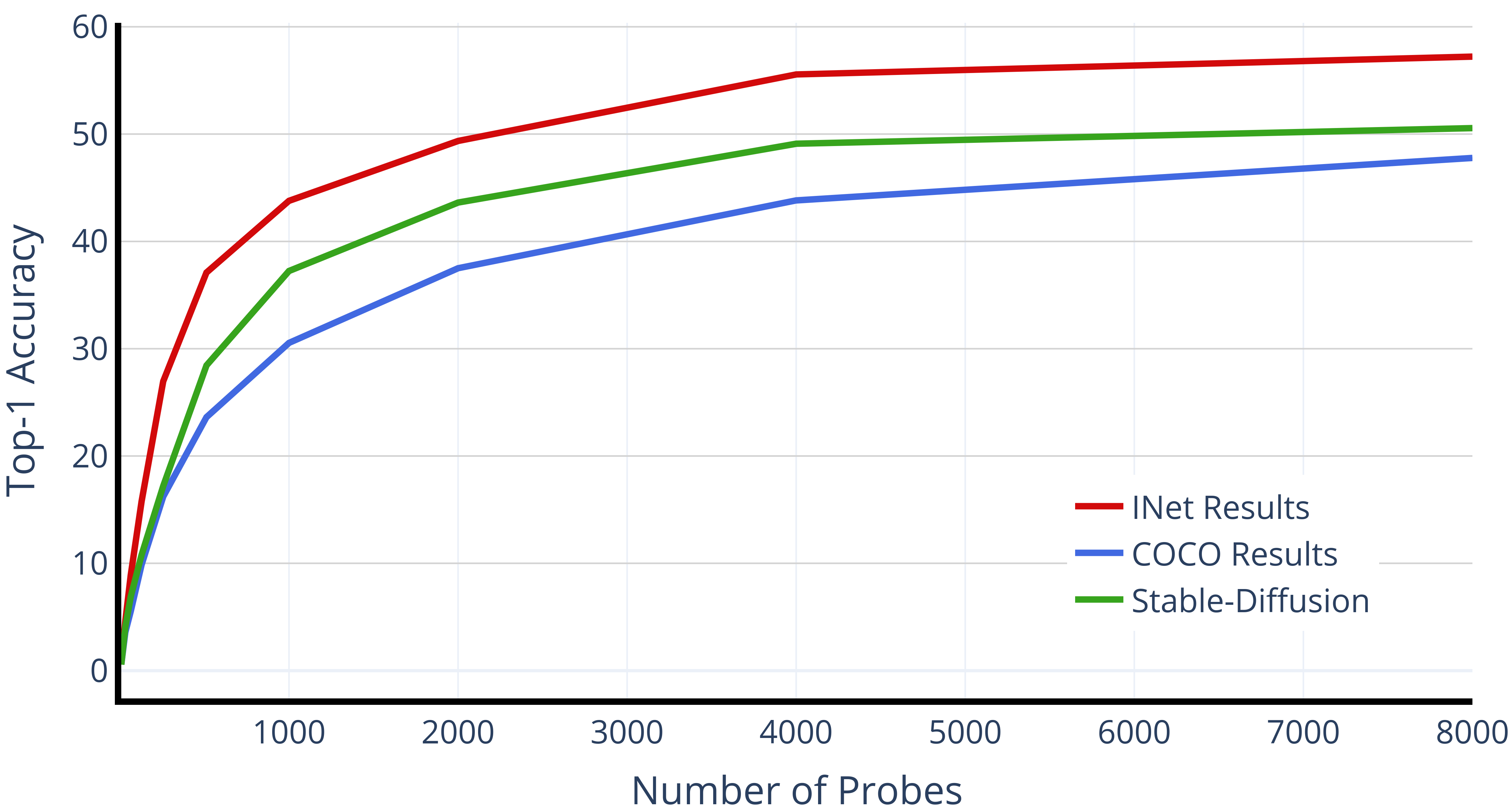}
    \caption{\textit{\textbf{Number of Probes.}} We test our zero-shot retrieval approach on INet-Hub with increasing numbers of probes. While more probes lead to higher accuracy, the gains are diminishing.}
    \label{fig:num_probes}
    \vspace{-3mm}
\end{figure}

\paragraph{How many probes are enough?} Fig.~\ref{fig:num_probes} presents results of text retrieval on INet-Hub using increasing numbers of probes. More probes lead to better results but with diminishing gains. For example, $4,000$ COCO probes are enough for good performance of $43.8\%$ top-1 accuracy, though it is possible to achieve a $47.8\%$ using $8,000$ probes.

\section{Discussion}

\paragraph{Non-random probe selection.} We proposed an approach for searching models that can recognize a target concept. Our approach probes each model with $4,000$ COCO images to produce the representation of each logit. However, we believe this number can be reduced substantially. For instance, while we chose the set of probes at random, it is likely that a smaller and more curated of probes exists. Specifically, core-set methods, which aim to reduce the number of training data, could potentially reduce this number. 

\paragraph{Scaling-up to entire repositories.} While our model hubs already have up to $1,500$ large models, including ViTs \citep{vit} and RegNet-Ys \cite{regnet}, large model repositories may contain a millions of models. We chose to test our approach on smaller hubs mainly because we did not have the resources to probe a million models. However, given ProbeLog representations for all models, the search should be fast and well within the capabilities of most researchers. The descriptors are lightweight compared to actual model weights, and storing them is quite cheap. E.g., our INet-Hub models require $400GB$ of memory, but their logit $8,000$ probes descriptors only consume $1.4GB$. Also, our search algorithm operates in a space of a few tens of dimensions, where retrieval from even a billion entries is possible \citep{johnson2019billion,jayaram2019diskann,chen2021spann}. 

\paragraph{Improved collaborative probing.} We showed that a simple collaborative filtering approach can significantly reduce the probing cost for repository. There are several ways to improve it. One direction is to develop a more sophisticated method for selecting which probes to sample for each model. This can be in an adaptive way i.e., sampling the first few prompts can inform the choice for the next  probes. Another direction is to use improved collaborative filtering ideas which take into account the statistics of logit values. We believe this is a fruitful avenue for future research.

\section{Limitations}

\paragraph{Extension beyond classification models.} Our proposed method embeds each logit of each model on its own. This will require modification for generative models where the output dimensions do not explicitly encode the learned concepts. While some works attempted to search for generative adapters \citep{contentsearch}, they typically required many more ($50,000$) probes as their descriptors summarize the distribution of probe outputs. We believe that our methodology, where the inputs are ordered and fixed for all models, can reduce the number of probes substantially.

\paragraph{Out-of-distribution concepts.} To enable search for diverse concepts we chose sampled probes from the COCO dataset \citep{coco} which does not just contain centered objects but also entire scenes. Still, this probe distribution does not represent all concepts, e.g. it does not include medical concepts. Successfully searching for such far OOD concepts will probably require selecting a probe distribution that is better aligned to the target concepts.

\section{Conclusion}
In this paper we propose an approach for searching for models in large repositories that can recognize a target concept. We first probe all models with a fixed, ordered set of probes, and define the values from each output dimension (logit) across all probes as a ProbeLog descriptor. We find that by normalizing these descriptors, we can compare them across different models, and even to zero-shot classifiers such as CLIP. With the pairwise discrepancy measure, we propose a method for searching models by text. We also present Collaborative Probing to significantly reduce the number of required probes at the same accuracy. We evaluate our approach on real-world models, and show it generalizes well to In-the-Wild models collected from HuggingFace. We ablated our design choices and showed they are crucial for effective model search.

\bibliography{example_paper}
\bibliographystyle{icml2025}

\newpage
\appendix
\onecolumn
\section{INet-Hub Dataset Details}
\label{app:inet_hub}
To simulate a model hub with many classifiers, we train $1,500$ classifier models on different subsets of ImageNet classes. Each classifier is trained on a subset of between $15$ and $200$ classes, where the classes are chosen at random separately for each model. $90\%$ of the classifiers are initialized from a foundation model, and the rest $10\%$ are trained from scratch. The pre-training weights are selected from a set of $49$ different models spanning various architectures including ViTs \citep{vit}, ResNets \citep{resnet}, RegNet-Ys \citep{regnet}, MLP Mixers \citep{mlp_mixer}, EfficientNets \citep{efficientnet}, ConvNexts \citep{convnext} and more. Each model is then trained for $2-5$ epochs. This process results in a model hub with over $85,000$ different logits to search for and $1,000$ different fine-grained concepts. Below we list the possible pre-training weights of each model. All pre-training weights are taken from the timm library \citep{rw2019timm}.

\begin{multicols}{2}
\begin{itemize}
    \item vit\_base\_patch32\_clip\_quickgelu\_224.laion400m\_e32
    \item vit\_base\_patch32\_clip\_224.laion400m\_e32
    \item vit\_base\_patch32\_clip\_224.laion2b
    \item vit\_base\_patch32\_clip\_224.datacompxl
    \item convnext\_base.clip\_laiona
    \item convnext\_base.clip\_laion2b
    \item vit\_base\_patch32\_clip\_quickgelu\_224.metaclip\_400m
    \item vit\_base\_patch32\_clip\_quickgelu\_224.metaclip\_2pt5b
    \item vit\_base\_patch32\_clip\_224.metaclip\_400m
    \item vit\_base\_patch32\_clip\_224.metaclip\_2pt5b
    \item vit\_base\_patch32\_clip\_224.openai
    \item seresnextaa101d\_32x8d.sw\_in12k
    \item resmlp\_24\_224.fb\_dino
    \item resmlp\_12\_224.fb\_dino
    \item mixer\_l16\_224.goog\_in21k
    \item mixer\_b16\_224.miil\_in21k
    \item mixer\_b16\_224.goog\_in21k
    \item resnetv2\_152x2\_bit.goog\_in21k
    \item resnetv2\_101x1\_bit.goog\_in21k
    \item resnetv2\_50x1\_bit.goog\_in21k
    \item regnety\_320.seer
    \item regnety\_160.sw\_in12k
    \item regnety\_120.sw\_in12k
    \item swin\_tiny\_patch4\_window7\_224.ms\_in22k
    \item swin\_base\_patch4\_window7\_224.ms\_in22k
    \item convnext\_small.in12k
    \item convnext\_tiny.in12k
    \item convnext\_tiny.fb\_in22k
    \item convnext\_small.fb\_in22k
    \item convnext\_nano.in12k
    \item convnext\_base.fb\_in22k
    \item eca\_nfnet\_l0
    \item vit\_base\_patch16\_224.dino
    \item vit\_small\_patch16\_224.dino
    \item vit\_base\_patch16\_224.mae
    \item vit\_base\_patch16\_224.orig\_in21k
    \item vit\_base\_patch32\_224.orig\_in21k
    \item vit\_tiny\_r\_s16\_p8\_224.augreg\_in21k
    \item vit\_small\_r26\_s32\_224.augreg\_in21k
    \item vit\_tiny\_patch16\_224.augreg\_in21k
    \item vit\_small\_patch32\_224.augreg\_in21k
    \item vit\_small\_patch16\_224.augreg\_in21k
    \item vit\_base\_patch32\_224.augreg\_in21k
    \item vit\_base\_patch16\_224\_miil.in21k
    \item vit\_base\_patch16\_224.augreg\_in21k
    \item tf\_efficientnetv2\_s.in21k
    \item tf\_efficientnetv2\_m.in21k
    \item tf\_efficientnetv2\_l.in21k
    \item tf\_efficientnetv2\_b3.in21k
\end{itemize}
\end{multicols}

\clearpage

\section{HF-Hub Dataset Details}
\label{app:hf_hub}
In order to test our method on real-world data, we collected $71$ classifiers uploaded by users to hugging face. Classifiers have between $2$ and $82$ classes they were trained on. Overall there are more than $400$ possible logits in the dataset. The models are trained on a diverse set of models, and class names are given by free text. Hence, class names may not align perfectly as each user spells concept a bit differently (e.g., ``Apple'' vs. ``Apples'). Moreover, some classifiers have different levels of granularity, such as ``Car'' vs. a specific car model ``Toyota''.
We therefore created a label mapping where we manually annotated to which classes each logit can be mapped. We follow these rules to allow mappings between labels: (i) Different spelling map to each other. (ii) An object can be mapped to a specific type of it, e.g. "cat" -> "siamese cat". (iii) a specific type of object can be mapped to its super-class e.g. "siamese cat" -> "cat". (iv) object of the same level of granularity that share a super class cannot be mapped to each other. For example, a "Golden Retriever" is not a good match for a "Husky". Additionally, we created an additional mapping which matches each class to its corresponding ImageNet concept when available.

\section{Additional Results}
\label{app:additional_results}

We present additional results with the baselines and the Top-5 precision metric as well, in Tab.\ref{tab:main_results_precision} 

\begin{table*}[h!]
    \caption{\textbf{\textit{Retrieval Results.}} We provide the additional Top-5 retrieval precisions of our method and the baselines, over several scenarios. All methods use COCO images as probes.}
    \vspace{0.2cm}
    \centering
    \begin{tabular}{c|cccccc}
    Retrieval & Method & INet $\rightarrow$ INet & INet $\rightarrow$ HF & HF $\rightarrow$ INet & text $\rightarrow$ HF & text $\rightarrow$ INet \\
    \toprule
    \multirow{2}{*}{Top-1} & Full Query & 59.9\%\std{0.2} & 14.8\%\std{0.1} & 15.3\%\std{0.8} & 22.6\%\std{0.5} & 16.9\%\std{0.2} \\
    \multirow{2}{*}{Accuracy} & Model-Level & 0\%\std{0.} & 13.9\%\std{1.0} & 21.0\%\std{1.8} & 17.8\%\std{1.5} & 0\%\std{0.0} \\
    & \textbf{Ours (ProbeLog)} & 72.8\%\std{0.2} & 26.1\%\std{0.8} & 40.6\%\std{0.3} & 34.0\%\std{1.5} & 43.8\%\std{1.1} \\
    \midrule
    \multirow{2}{*}{Top-5} & Full Query & 82.8\%\std{0.1} & 31.5\%\std{0.1} & 19.7\%\std{0.8} & 38.6\%\std{1.1} & 22.8\%\std{0.2} \\
    \multirow{2}{*}{Accuracy} & Model-Level & 0\%\std{0.} & 34.6\%\std{0.6} & 51.6\%\std{2.0} & 38.8\%\std{1.8} & 0\%\std{0.0} \\
    & \textbf{Ours (ProbeLog)} & 92.6\%\std{0.1} & 43.6\%\std{0.5} & 58.6\%\std{0.9} & 53.7\%\std{1.9} & 68.0\%\std{0.6} \\
    \midrule
    \multirow{2}{*}{Top-5} & Full Query & 61.2\%\std{0.1} & 9.7\%\std{0.1} & 13.6\%\std{0.3} & 13.1\%\std{0.3} & 14.1\%\std{0.1} \\
    \multirow{2}{*}{Precision} & Model-Level & 0\%\std{0.0} & 8.9\%\std{0.2} & 18.0\%\std{0.5} & 9.9\%\std{0.5} & 0\%\std{0.0} \\
    & \textbf{Ours (ProbeLog)} & 71.2\%\std{0.1} & 15.4\%\std{0.4} & 39.4\%\std{0.9} & 20.8\%\std{0.4} & 39.7\%\std{0.8} \\
    \end{tabular}
    \label{tab:main_results_precision}
\end{table*}


\end{document}